\documentclass[lettersize,journal]{IEEEtran}
\usepackage{amsmath,amsfonts}
\usepackage{algorithmic}
\usepackage{algorithm}
\usepackage{array}
\usepackage{booktabs}
\usepackage{multirow}
\usepackage{tabularx,ragged2e}
\newcolumntype{C}{>{\centering\arraybackslash}X}
\usepackage[caption=false,font=normalsize,labelfont=sf,textfont=sf]{subfig}
\usepackage{textcomp}
\usepackage{stfloats}
\usepackage{url}
\usepackage{verbatim}
\usepackage{graphicx}
\usepackage{cite}
\usepackage{hyperref}
\begin{document}

\title{On the Impact of Voice Anonymization on Speech Diagnostic Applications: a Case Study on COVID-19 Detection}

\author{Yi Zhu, Mohamed Imoussaïne-Aïkous, Carolyn Côté-Lussier, and Tiago H. Falk
\thanks{}
\thanks{The authors are with the Institut national de la recherche scientifique, University of Québec, Montréal, Canada.

Our code and voice demos are made available at \href{https://github.com/zhu00121/Anonymized-speech-diagnostics/}{https://github.com/zhu00121/Anonymized-speech-diagnostics}.}}



\maketitle

\begin{abstract}
With advances seen in deep learning, voice-based applications are burgeoning, ranging from personal assistants, affective computing, to remote disease diagnostics. As the voice contains both linguistic and para-linguistic information (e.g., vocal pitch, intonation, speech rate, loudness), there is growing interest in voice anonymization to preserve speaker privacy and identity. Voice privacy challenges have emerged over the last few years and focus has been placed on removing speaker identity while keeping linguistic content intact. For affective computing and disease monitoring applications, however, the para-linguistic content may be more critical. Unfortunately, the effects that anonymization may have on these systems are still largely unknown. In this paper, we fill this gap and focus on one particular health monitoring application: speech-based COVID-19 diagnosis. We test three anonymization methods and their impact on five different state-of-the-art COVID-19 diagnostic systems using three public datasets. We validate the effectiveness of the anonymization methods, compare their computational complexity, and quantify the impact across different testing scenarios for both within- and across-dataset conditions. Additionally, we provided a comprehensive evaluation of the importance of different speech aspects for diagnostics and showed how they are affected by different types of anonymizers. Lastly, we show the benefits of using anonymized external data as a data augmentation tool to help recover some of the COVID-19 diagnostic accuracy loss seen with anonymization.
\end{abstract}

\begin{IEEEkeywords}
Voice anonymization, health diagnostics, COVID-19 detection
\end{IEEEkeywords}

\section{Introduction}
\IEEEPARstart{S}{peech} is one of the most powerful and easy-to-use communication interfaces between humans and machines. For example, voice assistants relying on automatic speech recognition (ASR) allow humans to control devices by providing voice commands \cite{hoy2018alexa}; automatic speaker verification (ASV) systems enable users to access personal properties (e.g., online bank accounts) via their voice \cite{5871484}. More recently, speech has also been shown as a promising measure for in-home disease detection and monitoring, including COVID-19 \cite{lella2021literature}, chronic obstructive pulmonary disease (COPD) \cite{nathan2019assessment}, and Alzheimer's disease \cite{petti2020systematic}, just to name a few. 

Speech-based diagnostic systems have been motivated by the fact that speech requires complex and precise coordination of the respiratory system and neuromuscular control \cite{macneilage1980speech}. Diseases that cause dysfunction in speech production would then lead to changes in vocal characteristics. For example, major symptoms of COVID-19, such as cough, muscle soreness, and decreased neuromuscular control \cite{vetter2020clinical, quatieri2020framework, paliwal2020neuromuscular}, have been shown to relate to increased vocal hoarseness and variance in syllabic rate \cite{zhu2022fusion}. While human ears may not be able to capture such subtle changes, machine learning (ML) models have demonstrated the capability to detect certain abnormal patterns present in pathological speech \cite{zhu2022fusion, schuller2021interspeech, sharma2022second}.

Today, the great majority of speech-based applications rely on deep neural network (DNN) architectures with models containing hundreds of millions of parameters, with this number continuously rising. Commonly, these parameters are not stored locally on mobile devices \cite{wang2018deep} and speech data are sent to and processed in the cloud; decisions are then transmitted back to the user device. As more and more cases of cyberattacks are being reported \cite{stupp2019fraudsters, kaloudi2020ai, yamin2021weaponized}, this transmission of speech data over the cloud could pose serious threats to user privacy. It has been previously reported that voice assistants and many third-party applications collect users' voices without their knowledge and share it with advertising partners \cite{iqbal2022your}. For example, Amazon patented a technique which recognizes health status via conversations with users and advertises the related medicines to them \cite{jin2018voice}. This could be particularly risky for speech diagnostics applications, since the user's voice could be linked with sensitive medical information, such as health status\cite{latif2020speech}, disease progression\cite{harel2004acoustic}, or mental state \cite{low2020automated}, just to name a few. As such, speech privacy-preserving methods have gained increased attention globally, especially with the release of regulations, such as the General Data Protection Regulation (GDPR) in Europe \cite{zarsky2016incompatible} and the Personal Information Protection Law (PIPL) in China \cite{calzada2022citizens}; the latter is particularly aimed at personal biometrics (i.e., voice, facial image, and fingerprints).

Alternately, voice anonymization methods have emerged with the aim of manipulating the speech signal such that information about speaker identity is obfuscated, while the linguistic content and other para-linguistic attributes (e.g., timbre, naturalness) remain intact. Given the burgeoning interest in this domain, the Voice Privacy Challenges (VPC) were held in 2020 and 2022 to foster development in speech anonymization techniques \cite{tomashenko2020voiceprivacy, tomashenko2022voiceprivacy}. However, these challenges were aimed at developing anonymization methods for downstream automatic speech recognition tasks \cite{tomashenko2020voiceprivacy, tomashenko2022voiceprivacy, fang2019speaker, meyer2022speaker}, where linguistic content was preserved, but not para-linguistic information. 

As speech applications emerge beyond the realm of ASR, it is important to also gauge what impacts anonymization tools can have on other downstream tasks. Some initial attempts have been made in this realm. Nourtel et al. showed significant degradation in speech emotion recognition when anonymization was applied \cite{nourtel2021evaluation}. Dumpala et al. performed an initial exploration of the privacy-preserving features of depression speech \cite{dumpala2021sine}. To the best of our knowledge, gauging the impact of anonymization on speech diagnostic applications has yet to be explored; this paper aims to fill that gap. 

Furthermore, in a real-world scenario, diagnostics models are usually trained on open-source datasets due to the scarcity of medical data \cite{lee2017medical}, while test data may come from varying conditions (e.g., geographic locations, languages, collection devices, etc.). Hence, it is difficult to have training and test data anonymized using the exact same approach in reality. However, existing anonymization testing conditions commonly assume that downstream models either have no or full knowledge of how the training and test data are anonymized (i.e., \textit{ignorant} or \textit{fully-informed}, respectively). As more anonymization techniques emerge, alternate testing conditions could be implemented, such as training with data processed by other anonymization methods (i.e., in a \textit{semi-informed} manner) or with both original and conventional anonymization tools (i.e., \textit{augmented}). Hence, more complex testing conditions need to be considered. Lastly, to avoid private information being sent to the cloud, the voice anonymization should be deployed locally on the user device, which could have limited computational resources. As such, it is important to evaluate the computational complexity (i.e., time and capacity needed for computation) of the anonymization methods alongside their effectiveness.


In this study, we comprehensively evaluated the impact of three voice anonymization methods on the accuracy of five leading COVID-19 detection systems. We started by quantifying the efficacy and computational complexity of the anonymization methods with COVID-19 speech recordings. We then investigated the within and cross-dataset performance of five COVID-19 diagnostics systems in different conditions, and explored the reasons behind the impact of different anonymization methods on diagnostics. Lastly, we showed the benefits of using anonymized external data as a data augmentation tool to recover the diagnostics accuracy loss in anonymized data. The following paper is organized as follows. Section II summarizes the related works in speech-based COVID-19 diagnostics and speech anonymization. Section III and IV describe the main components of the anonymized speech diagnostics framework and the experimental set-up. Section V describes and discusses the obtained results. Section VI presents the conclusions.
\section{Related Work}

\subsection{Speech-based COVID-19 Diagnostics}
Speech-based diagnostic systems can be categorized into two groups: ones that rely on carefully designed hand-crafted features coupled with conventional machine learning classifiers, and ones that input raw signals directly into a deep learning model for classification. In the latter `end-to-end' scenario, the deep learning model serves as a feature extractor and feature mapping function in one. 

When it comes to feature extraction from speech, the openSMILE toolkit  \cite{eyben2010opensmile} is by far the most popular. The largest feature set of openSMILE extracts over 6,000 acoustic features, including mel-frequency cepstral coefficients (MFCC), pitch contours, voicing-related information, as well as several other low-level descriptors (LLDs). This feature set has been used together with conventional classifiers, such as support vector machines (SVM), for the detection of different diseases \cite{warnita2018detecting, nallanthighal2022detection, han2020early,han2022sounds}. More recently, it has been employed as a benchmark feature set for the INTERSPEECH 2021 ComParE COVID-19 Detection Challenge \cite{coppock2022summary}. For in-the-wild speech analysis, on the other hand, the modulation spectral representation (MSR) has shown benefits over openSMILE features for different applications (e.g., \cite{falk2009modulation, avila2018feature}), including disease characterization (e.g., \cite{tiwari2022modulation,falk2010spectro}) and COVID-19 detection \cite{zhu2022fusion}. 


Existing end-to-end systems, in turn, have relied on variants of the spectrogram representation as input, including the mel-spectrogram or the log-mel-spectrogram, as well as convolutional or recurrent neural network architectures for classification. Han et al., for example, showed that VGGish neural networks outperformed conventional methods in classifying different COVID-19 symptoms \cite{han2022sounds}. Akman et al. developed a ResNet-like architecture for speech and cough-based COVID-19 detection \cite{akman2021evaluating}. The Bi-directional Long-Short-Term-Memory (BiLSTM) neural network was used in the top-performing system competing in the second Diagnosis of COVID-19 using Acoustics (DiCOVA2) Challenge \cite{sharma2022second}. Compared to conventional systems, end-to-end systems have demonstrated overall higher performance on several datasets without the need for a separate feature extraction step \cite{deshpande2020audio, schuller2021covid, sharma2022second}. Nonetheless, recent research has shown that while end-to-end models achieve state-of-the-art accuracy on a particular dataset, those results do not transfer well to other unseen datasets, where accuracy can drop to below chance levels \cite{zhu2022generalizable}; this was not the case with hand-crafted features and conventional classifiers.

\subsection{Speech Anonymization}
Anonymization techniques comprise two categories: speech transformation and speech conversion. The former refers to modifications directly to the original speech, such as pitch shifting and warping \cite{stylianou2009voice, srivastava2020evaluating}, to remove personal identifiable information from the speech signal. The latter, in turn, converts one's voice to sound like that of another without changes in linguistic content \cite{mohammadi2017overview}. As voice privacy concerns are on the rise, voice anonymization has gained popularity recently and, in 2020, the Voice Privacy Challenge (VPC) was created \cite{tomashenko2020voiceprivacy}. A popular method from the 2020 and 2022 VPCs employs the so-called McAdams coefficients \cite{tomashenko2020voiceprivacy, tomashenko2022voiceprivacy}, where shifts in the pole positions derived from linear predictive coding (LPC) analysis of speech signals \cite{mcadams1984spectral} are used to achieve anonymization. Another popular voice transformation method is termed voicemask \cite{qian2017voicemask}, where certain frequency components are compressed (or stretched) to generate a lower-pitched (or higher-pitched) voice signal. Voice conversion systems, on the other hand, have usually relied on modifications to speaker embeddings, such as the x-vector \cite{snyder2018x} and the ECAPA-TDNN embeddings \cite{desplanques2020ecapa}, which are assumed to only carry nonverbal information that pertains to the speaker identity alone. The modified speaker embeddings are then input with speech content sequence to a speech synthesis module to reconstruct a new speech waveform \cite{fang2019speaker}. Several innovations have been proposed to the speech synthesis module to make the outcome sound more natural and of greater quality and intelligibility \cite{kong2020hifi, srivastava2020design, meyer22b_interspeech, meyer2022speaker}.

\section{Anonymized Speech Diagnostics Systems}

\subsection{System Overview}
\begin{figure}[b]
  \includegraphics[width=.95\linewidth]{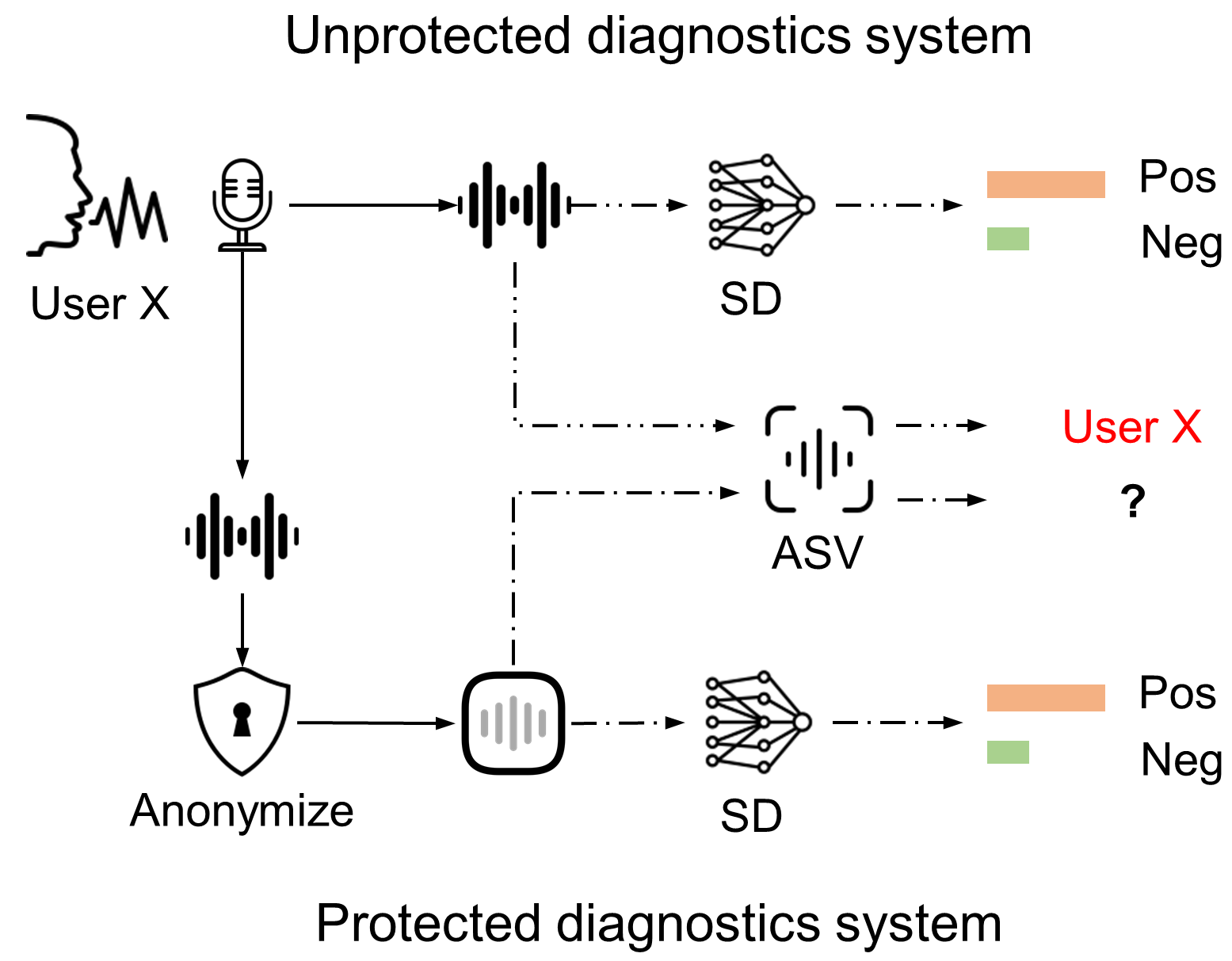}
  \caption{Block diagram of a speech-based diagnostics system with (protected) and without (unprotected) anonymization. `SD' stands for speech-based diagnostic system and `ASV' for automatic speaker verification.}
  \label{unprotected}
\end{figure}

Figure~\ref{unprotected} depicts the diagram of an anonymized speech diagnostics (SD) system. Conventionally, the original voice of user X is input to a diagnostic system that will generate a positive or negative output for the tested disease and/or symptom. If an automatic speaker verification (ASV) system was trained with data from user X, the ASV system would be able to detect user X's voice. In practice, SD systems are complex and models are often stored on the cloud, thus requiring the user's voice (or features) to be uploaded to the cloud. This transmission of data could result in privacy concerns. To overcome this, voice anonymization can be employed locally and anonymized data (or features) are sent to the cloud. In this case, user X would not be identified by the ASV system and speech-based diagnostics could proceed in a more secure and private manner. 

%

\subsection{Speech-Based Diagnostic Systems}
Based on previous experiments on COVID-19 detection (e.g., \cite{zhu2022fusion}), the five top-performing diagnostics systems are explored herein:\\
\subsubsection{openSMILE+SVM} A total of 6,373 static acoustics features were firstly extracted using the openSMILE toolbox \cite{eyben2010opensmile}, which were then input to a SVM classifier with a linear kernel. This system was used as the benchmark in the 2021 ComParE COVID-19 Speech Sub-challenge \cite{coppock2022summary}.
\subsubsection{openSMILE+PCA+SVM} The high dimensionality of the openSMILE features can be problematic for smaller datasets. In \cite{xia2021covid}, principal component analysis (PCA) \cite{wold1987principal} was used to compress the 6,000+ features into 300 components. Here, the number of principal components was treated as a hyper-parameter and a value of 100 was found to strike a good balance in accuracy and dimensionality. 
\subsubsection{MSR+SVM} The MSR features have been used in \cite{zhu2022fusion, zhu2022generalizable} and shown to outperform openSMILE-based systems and to provide improved generalizability across datasets. The interested reader is referred to \cite{tiwari2022modulation,santos2014improved} for more details about the modulation spectrum. The modulation spectrum decomposes each frequency component along time into different modulation frequencies, which captures the abnormalities in respiration and articulation by focusing on long-term dynamics of speech. Each modulation spectrum comprises 23 frequency bins and 8 modulation frequency bins, which is then flattened into a vector and used as input to a linear SVM classifier.
\subsubsection{MSR+PCA+SVM} For more direct comparisons with the openSMILE system, here we also explore the compression of the 184-dimensional ($23\times8$) vector via PCA, resulting in a final 100-dimensional vector for classification.
\subsubsection{Logmelspec+BiLSTM} 
The winning system in the DiCOVA2 Challenge was employed~\cite{sharma2022second} as a benchmark. This system adopts the conventional log-mel-spectrogram (logmelspec) with first-and second-order deltas as input, along with a BiLSTM as the classifier. More details about the network architecture can be found in~\cite{sharma2022second}.

\subsection{Speech Anonymization Methods}
A voice transformation and two voice conversion methods are explored here to gauge their differences in speech diagnostics performance. More details are provided below. 

\subsubsection{McAdams coefficient} This approach uses a classical signal processing technique and does not require model training. It employs the so-called McAdams coefficient method \cite{patino2021speaker,mcadams1984spectral} to shift the position of formants measured using linear predictive coding (LPC) \cite{o1988linear}. For each short-time speech frame, the method first separates the linear prediction residuals and linear prediction (LP) coefficients. The LP coefficients are then converted to pole positions in the z-plane by polynomial root-finding, where each pole position represents the position of one formant. The phase of the poles with imaginary parts is then raised to the power of the McAdams coefficient $\alpha$. The new set of poles is then converted back to LP coefficients. Together with the original residuals, a new speech frame can be synthesized. 


\subsubsection{Ling-GAN} 
\begin{figure}
  \includegraphics[width=\linewidth]{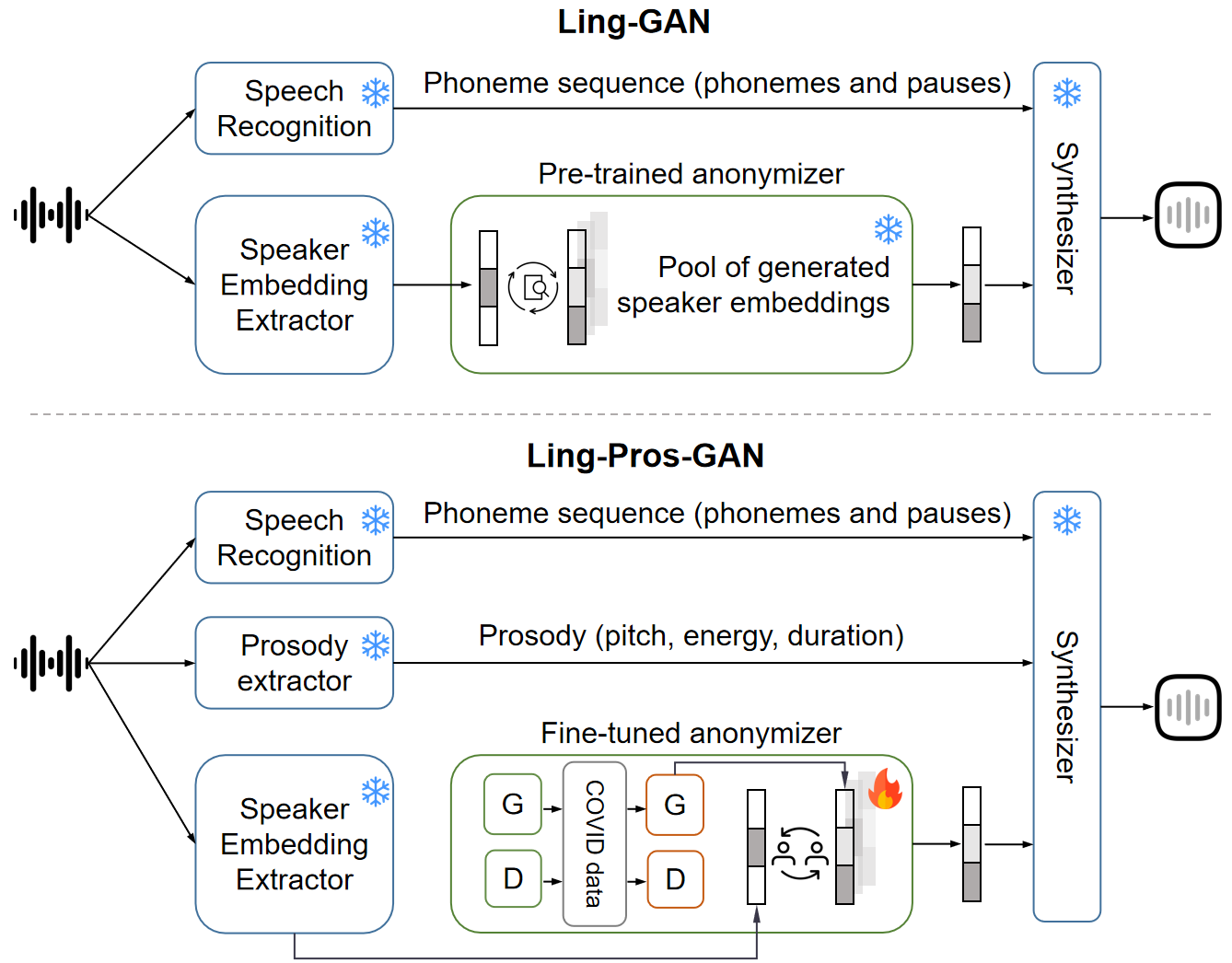}
  \caption{Diagram of the two GAN-based anonymizers implemented in this study. Compared to the Ling-GAN, the Ling-Pros-GAN not only preserves the original prosody, but also has the generator and discriminator fine-tuned with COVID-19 speech data, enabling it to generate more COVID-like speaker embeddings.}
  \label{gan}
\end{figure}

For voice conversion, we implemented two systems based on generative adversarial networks (GAN). The overall architecture of these systems can be found in Figure~\ref{gan}. The first system, abbreviated as `Ling-GAN', was an off-the-shelf anonymizer from \cite{meyer2022speaker}, where all modules were already trained and applied to COVID-19 data without any fine-tuning. In general, it preserves the linguistic content (i.e., phoneme sequence) and uses a generator to generate fake, yet realistic speaker embeddings to substitute the original speaker embeddings. The original speech is first input to an automatic speech recognition (ASR) model to extract the phone sequence. The ASR model used here is based on the hybrid CTC (Connectionist Temporal Classification)/attention architecture \cite{watanabe2017hybrid} with a Conformer encoder \cite{gulati2020conformer} and a Transformer decoder. It should be emphasized that the output of the ASR is a phoneme sequence, detailing not only the phonemes uttered but also the pauses. In our exploratory analysis, we found that the removal of these pauses would change the rhythm of the generated speech and lead to degraded diagnostic performance. We hence kept all pauses in the extracted phoneme sequences. The ASR model used here supports English as the default input language, hence may lead to erroneous transcriptions when other languages are used. Although such issue can be potentially tackled by replacing with other multi-language ASR models, their compatibility with the anonymization and synthesizer blocks has not been tested. Hence, we remain using the same architecture as is proposed in \cite{meyer2022speaker}, and leave the language compatibility for future investigation. 

The anonymization is divided into two stages. During the first stage, the 512-dimensional x-vector \cite{snyder2018x} and the 192-dimensional ECAPA-TDNN vector \cite{desplanques2020ecapa} are extracted using the SpeechBrain toolkit \cite{ravanelli2021speechbrain} and concatenated as the final speaker embeddings. At the second stage, a Wasserstein GAN with Quadratic Transport Cost (WGAN-QC) \cite{liu2019wasserstein} is used to generate a pool of 5,000 `converted' speaker embeddings and saved for later use.  When a new recording is input to the system, the model iteratively looks through the pool, and stops when it finds one with a cosine distance above 0.3 with the original speaker embeddings. This set of new embeddings are then used to substitute the original one for synthesis. The 0.3 threshold value of cosine distance was suggested from \cite{meyer2022speaker}, which ensures sufficient difference in speaker traits while maintaining the naturalness. Finally, the FastSpeech 2 model \cite{ren2020fastspeech} is used to synthesize the phone sequence into a spectrogram, followed by a HiFiGAN vocoder \cite{kong2020hifi} to convert the spectrogram into a final speech waveform. The synthesizer is conditioned on the anonymized speaker embedding, hence keeping the linguistic content while obfuscating the speaker identity.

It is important to emphasize that this off-the-shelf GAN has not seen pathological speech data during its training \cite{meyer2023prosody}. As a consequence, the generated speaker embeddings may not encapsulate health-related attributes, thus affecting diagnostic accuracy. The last anonymization system used overcomes this limitation, as detailed next. 
\subsubsection{Ling-Pros-GAN}
The second GAN-based system, abbreviated as `Ling-Pros-GAN', was modified from \cite{meyer2023prosody} which can be seen as a more advanced version of the Ling-GAN. While sharing similar architecture, such as the ASR module and the synthesizer, the Ling-Pros-GAN further preserves prosody (i.e., pitch, energy, and duration) during anonymization and uses the style embeddings from \cite{wang2018style} to represent speaker attributes. In addition, we fine-tuned the generator and discriminator using the aggregated training set data from all three COVID-19 datasets employed in this study. The goal of fine-tuning was to enable the GAN to generate COVID-like speaker embeddings. 

\begin{figure}
\centering
\includegraphics[width=0.9\linewidth]{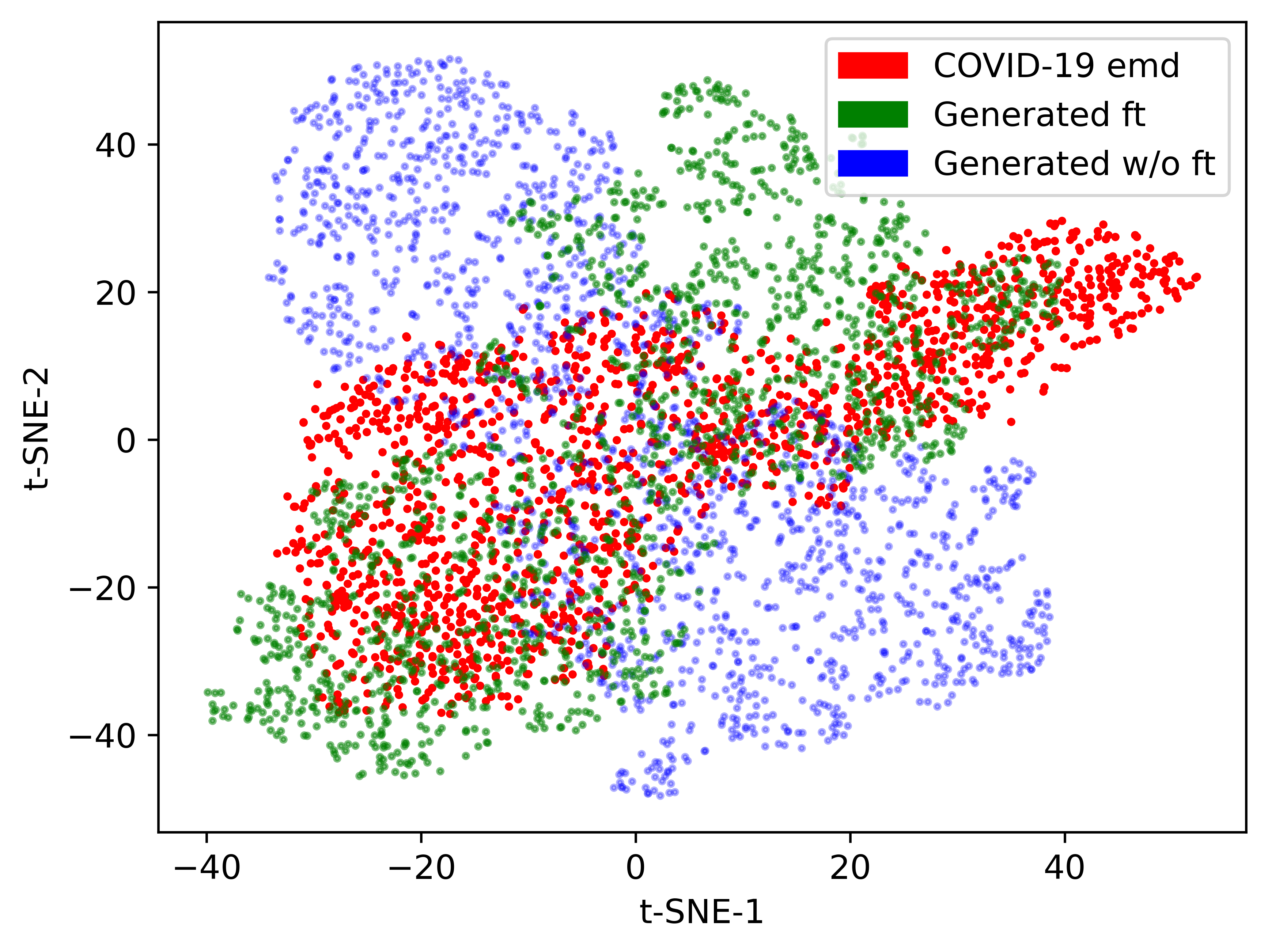}
\caption{Distribution of speaker embeddings (`emd') generated by Ling-Pros-GAN with and without fine-tuning (`ft'). Embeddings are projected to the 2-dimensional space using t-SNE.}
\label{fig:gan2}
\end{figure}

The generator and discriminator were jointly trained via 2,000 iterations, with the batch size of 128 and learning rate of .00005. Other fine-tuning hyperparameters remained the same as reported in \cite{meyer2023prosody}, which can also be found in our code repository. Figure~\ref{fig:gan2} depicts the t-distributed stochastic neighbor embedding (t-SNE) plots \cite{van2008visualizing} showing 
a 2-dimensional representation of the speaker embeddings in the COVID-19 datasets (red dots), those produced by the generator without fine-tuning (blue), and after fine-tuning (green). As can be seen, using just the pre-trained generator is not sufficient to model the COVID-19 speaker embedding distribution. With 2,000 iterations of fine-tuning, the generator was able to generate embeddings following a similar distribution of the COVID-19 embeddings. 

Different from the original implementation in \cite{meyer2023prosody}, where a pre-generated pool of speaker embeddings were used, we modified Ling-Pros-GAN in a way that it randomly generates a small set of different speaker embeddings each time it receives a new recording, then chooses which embeddings to swap by iteratively examining the cosine similarity. In other words, Ling-Pros-GAN is guaranteed to generate an unseen version of anonymized speech even with the exact same input recording. In contrary, since Ling-GAN always chooses embeddings from a pre-generated pool, there is a slight chance that two recordings may be anonymized with the same generated embeddings. Such possibility becomes higher when the number of speakers increases. While such modification to Ling-pros-GAN improves the privacy, the computing time increases simultaneously due to the online generation process of speaker embeddings.

\section{Experimental Setup}
\subsection{Databases}
At the time of writing, most existing COVID-19 sound datasets target cough sound, such as the COUGHVID \cite{orlandic2021coughvid}, Tos COVID-19 \cite{pizzo2021iatos}, Virufy \cite{chaudhari2020virufy}, and NoCoCoDa \cite{cohen2020novel}. Speech sound, on the other hand, is included in fewer datasets. To maximize the variability of data distribution and avoid biased results from one single dataset, we included three publicly available COVID-19 speech datasets, namely the multilingual 2021 ComParE COVID-19 Speech Sub-challenge (CSS) dataset~\cite{schuller2021interspeech}, the second DiCOVA Challenge dataset~\cite{sharma2022second}, and the English subset from the Cambridge COVID-19 sound database~\cite{xia2021covid}. These datasets are referred to hereinafter as CSS, DiCOVA2, and Cambridge set, respectively. The demographics of three datasets are summarized in Table~\ref{tab:data}. It should be noted that though the full Cambridge database contains more speech samples, the English subset has been more carefully examined by the data holders to avoid potential confounding factors (e.g., languages, data quality, class balance, etc.) \cite{xia2021covid}, hence is considered more suitable for our analysis. 

All three datasets were crowdsourced, volunteers across the globe were encouraged to upload their voice data and metadata via apps. The same speech content was required per dataset. With CSS, participants were asked to utter the sentence ``I hope my data can help to manage the virus pandemic'' at most three times in their mother tongue, with the majority of samples being uttered in English, Portuguese, Italian, and Spanish. The same speech content was used for the Cambridge set but in English only. With DiCOVA2, participants did number counting from 1 to 10 at a normal pace in English. For all datasets, participants were asked to self-declare whether they were COVID-negative (including healthy or having COVID-like symptoms) or COVID-positive (including symptomatic and asymptomatic cases). It can be noticed from Table~\ref{tab:data} that all three sets contained 10\% to 30\% asymptomatic COVID-positive cases. Additionally, nearly half of the COVID-negative samples in CSS and Cambridge are symptomatic, which is three times higher than that in DiCOVA2.

The CSS and Cambridge datasets were partitioned into three separate subsets by the challenge organizers, namely training, validation, and test. For comparisons, we employed the same challenge partition in this study. It should be emphasized that in the CSS dataset, several COVID-positive recordings were originally sampled at 8~kHz while the majority of the other files were sampled at 16~kHz. As suggested in \cite{coppock2022summary} and our previous exploration \cite{zhu2022fusion}, keeping these up-sampled recordings has been shown to lead to overly-optimistic results since classifiers learned to capture the difference in sampling rates instead of the actual pathological pattern. Thus, we removed them from our analysis. The DiCOVA2 dataset, in turn, is comprised of development and evaluation subsets, with the evaluation data being accessible only to challenge participants. Hence, we performed a speaker-independent training-test split (80/20\%) using the development subset only and left the evaluation set for testing. 
%
\begin{table*}
\caption{Dataset description and partitions. P-s: Symptomatic COVID-positive. P-a: Asymptomatic COVID-positive. N-s: Symptomatic COVID-negative. N-a: Asymptomatic COVID-negative. N/A: Information not provided.}
  \begin{tabular}{cccccccccccccccc}
    \hline
    \multirow{2}{*}{Dataset} & \multirow{2}{*}{Duration} & \multirow{2}{*}{Language} & \multicolumn{4}{c}{Symptomatic ratio} & \multicolumn{2}{c}{Gender} & \multicolumn{3}{c}{Age} & \multirow{2}{*}{Partition} & \multicolumn{2}{c}{COVID-label} &  \multirow{2}{*}{Total}  \\ 
    \cmidrule(lr){4-7}
    \cmidrule(lr){8-9}
    \cmidrule(lr){10-12}
    \cmidrule(lr){14-15}
    & & & P-s & P-a & N-s & N-a & Male & Female & $\leq$30 & 30-60 & $\geq$60 & & Pos & Neg & \\
    \\ \hline
    \multirow{3}{*}{CSS} & \multirow{3}{*}{3.24 hrs} & \multirow{3}{*}{Multi} & \multirow{3}{*}{72\%} & \multirow{3}{*}{28\%} & \multirow{3}{*}{41\%} & \multirow{3}{*}{59\%} & \multirow{3}{*}{56\%} & \multirow{3}{*}{43\%} & \multirow{3}{*}{11\%} & \multirow{3}{*}{70\%} & \multirow{3}{*}{19\%} & train & 56 & 243 & 299 \\
    & & & & & & & & & & & & valid & 130 & 153 & 283 \\
    & & & & & & & & & & & & test & 87 & 189 & 266 \\
    \hline
    \multirow{2}{*}{DiCOVA2} & \multirow{2}{*}{3.93 hrs} & \multirow{2}{*}{EN} & \multirow{2}{*}{87\%} & \multirow{2}{*}{13\%} & \multirow{2}{*}{14\%} & \multirow{2}{*}{86\%} & \multirow{2}{*}{74\%} & \multirow{2}{*}{26\%} & \multirow{2}{*}{43\%} & \multirow{2}{*}{52\%} & \multirow{2}{*}{5\%} &  develop & 137 & 635 & 772 \\
    & & & & & & & & &  & & & evaluate & 35 & 158 & 193\\
    \hline
    \multirow{3}{*}{Cambridge} & \multirow{3}{*}{5.29 hrs} & \multirow{3}{*}{EN} & \multirow{3}{*}{87\%} & \multirow{3}{*}{13\%} & \multirow{3}{*}{47\%} & \multirow{3}{*}{54\%} & \multirow{3}{*}{50\%} & \multirow{3}{*}{50\%} & \multirow{3}{*}{24\%} & \multirow{3}{*}{56\%} & \multirow{3}{*}{20\%} & train & 490 & 530 & 1020 \\
    & & & & & & & & & & & & valid & 82 & 60 & 142 \\ 
    & & & & & & & & & & & &  test & 162 & 162 & 324 \\ 
    \hline
  \end{tabular}
  \label{tab:data}
\end{table*}

\subsection{Tasks}
As our final goal is to not only provide accurate diagnostics decisions but also ensure the protection of privacy of speaker identity, the evaluation was divided into three tasks. In the first task, we compared the effectiveness and complexity of different anonymization techniques. In Task-2, we then quantified the impact of anonymization techniques on diagnostics accuracy in different conditions. Finally, we provided explanations for the impact seen in Task-2, and explored solutions for improving the proposed systems.

\subsubsection{Task-1: Evaluating anonymization performance}
\begin{figure}
  \includegraphics[width=\linewidth]{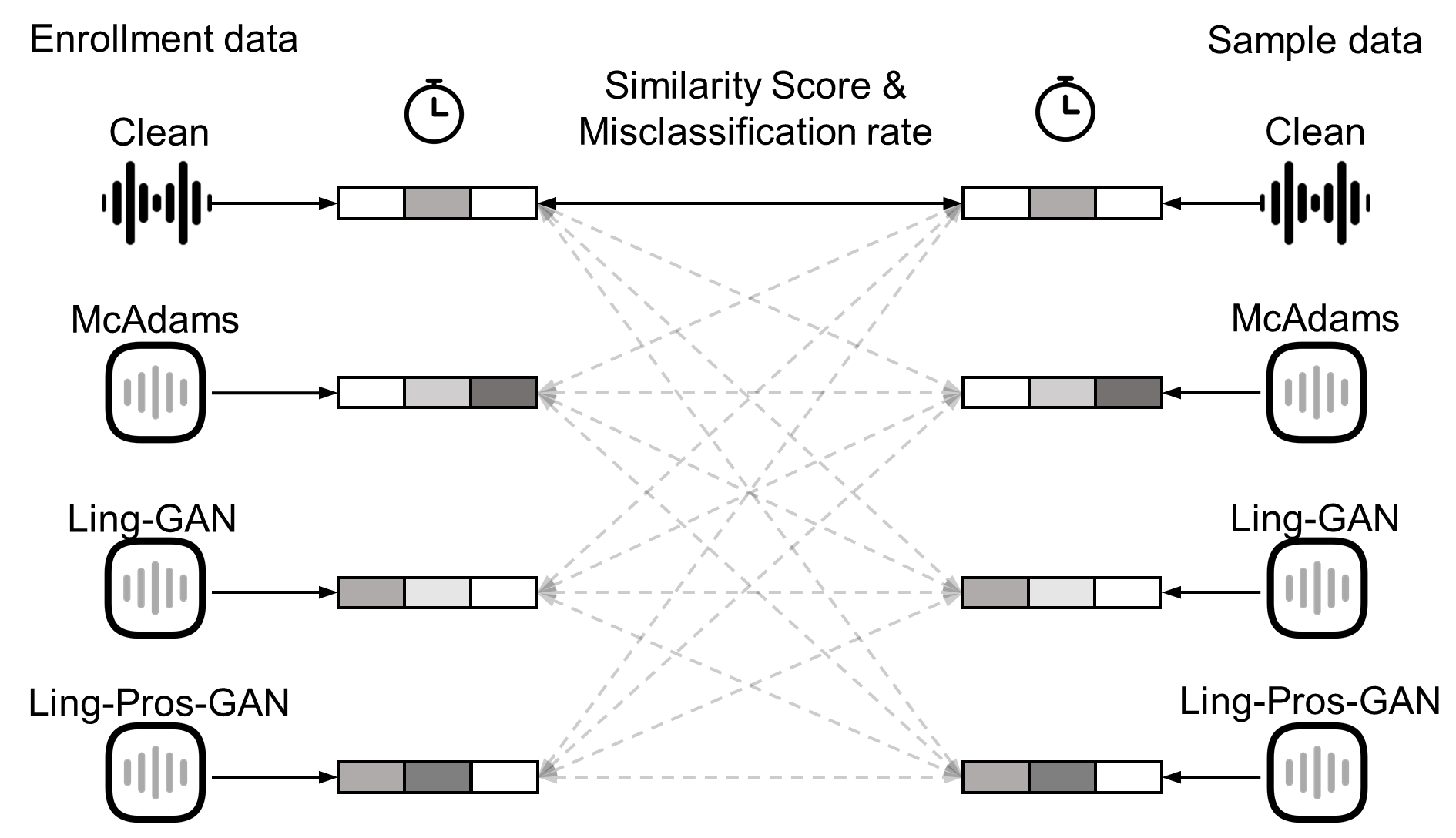}
  \caption{Evaluation of the effectiveness of different voice anonymization methods, as well as their computational complexity.}
  \label{asv}
\end{figure}

As is shown in Figure~\ref{asv}, for each speech recording, the speaker embeddings were extracted separately from the original version, the McAdams-anonymized version, the Ling-GAN anonymized version, and the Ling-Pros-GAN anonymized version. Cosine similarity was then computed between the embeddings of each two signals, where higher cosine similarity values represented higher resemblance between two speech samples. Meanwhile, we employed the pre-trained ECAPA-TDDN speaker verification model from SpeechBrain \cite{ravanelli2021speechbrain} to detect if two recordings are from the same speaker, then evaluated the misclassification rate, where higher values suggest more successful anonymization. Since multiple evaluation scenarios were considered in this study, where training and test data were processed with different anonymization methods, the cosine similarity and the misclassification rate were computed between not only the clean and anonymized data, but also data processed by different anonymization methods. Additionally, we measured the computation time spent by the three methods per recording, and calculated the average and standard deviation for each dataset. This helps to quantify and compare the time efficiency of the three anonymization methods.

\subsubsection{Task-2: Evaluating diagnostics accuracy}
As aforementioned in Section I, training and test data could be anonymized using different methods. To mimic a realistic setting, we explore four different scenarios, as detailed below. Table~\ref{tab:lookup} summarizes these conditions.\\
\textbf{Scenario-A: Unprotected:} Here, both training and test data are original, thus anonymization is not performed. This encompasses the traditional diagnostic system evaluation and serves as a baseline of the maximum diagnostics accuracy that can be achieved by each model.\\
\textbf{Scenario-B: (Anonymization) Ignorant:} In this scenario, the training data are original, and only the test data are anonymized. This scenario can be further separated into three cases: test data are anonymized using the McAdams coefficient (scenario \textbf{B1}), the Ling-GAN (scenario \textbf{B2}), and the Ling-Pros-GAN (scenario \textbf{B3}). This scenario exemplifies the case where new anonymization methods are proposed and tested against legacy original diagnostic systems.\\
\textbf{Scenario-C: Semi-informed:} In this scenario, anonymized data are seen during training, but from a method different from that used for testing. Six combinations were possible out of the three systems, namely: training set comprised of McAdams coefficient anonymization and test set with Ling-GAN (\textbf{C1}), training set with McAdams anonymizer and test set with Ling-Pros-GAN (\textbf{C2}), training set with Ling-GAN and test set with McAdams anonymizer (\textbf{C3}), training set with Ling-GAN and test with Ling-Pros-GAN (\textbf{C4}), training set with Ling-Pros-GAN and test with McAdams anonymizer (\textbf{C5}), and training set with Ling-Pros-GAN and test with Ling-GAN (\textbf{C6}). This scenario exemplifies the case where new anonymization methods are proposed and tested against legacy or different anonymized systems.\\
\textbf{Scenario-D: Fully-informed:} In this setting, training and test data are both anonymized using the same method and parameters, with three cases: both are anonymized with the McAdams coefficient (\textbf{D1}) method, both with Ling-GAN (\textbf{D2}), and both with Ling-Pros-GAN (\textbf{D3}).

\begin{table}[tb]
\caption{Training/test set details for the different conditions and scenarios explored.}
\centering
\begin{tabular}{cccc}
\toprule
Scenarios & Sub-condition & Training anonym. & Test anonym. \\
\midrule \midrule
Unprotected & A & Clean & Clean \\
\midrule
\multirow{3}{*}{Ignorant} & B1 & Clean & McAdams Coefs \\
& B2 & Clean & Ling-GAN \\
& B3 & Clean & Ling-Pros-GAN \\
\midrule
\multirow{6}{*}{Semi-informed} & C1 & McAdams Coef & Ling-GAN \\
& C2 & McAdams Coef & Ling-Pros-GAN \\
& C3 & Ling-GAN & McAdams Coefs \\
& C4 & Ling-GAN & Ling-Pros-GAN \\
& C5 & Ling-Pros-GAN & McAdams Coefs \\
& C6 & Ling-Pros-GAN & Ling-GAN \\
\midrule
\multirow{3}{*}{Fully-informed} & D1 & McAdams Coefs & McAdams Coefs\\
& D2 & Ling-GAN & Ling-GAN \\
& D3 & Ling-Pros-GAN & Ling-Pros-GAN \\
\bottomrule
\end{tabular}
\label{tab:lookup}
\end{table}

As data distributions vary across datasets \cite{roberts2021common}, diagnostics performance obtained under within-dataset conditions may lack external validity and has been shown to be over-optimistic \cite{roberts2021common, akman2021evaluating, zhu2022generalizable}. To ensure the generalizability of the tested methods, for each scenario we explore both within- and cross-dataset results. In the latter, models are trained on one dataset and tested on data from another set. As the CSS is a subset of the Cambridge set, we avoid using both datasets in the cross-database condition to avoid overly-optimistic results \cite{xia2021covid, coppock2022summary}.

\subsubsection{Task-3: Anonymization for data augmentation}
\begin{figure}
  \includegraphics[width=\linewidth]{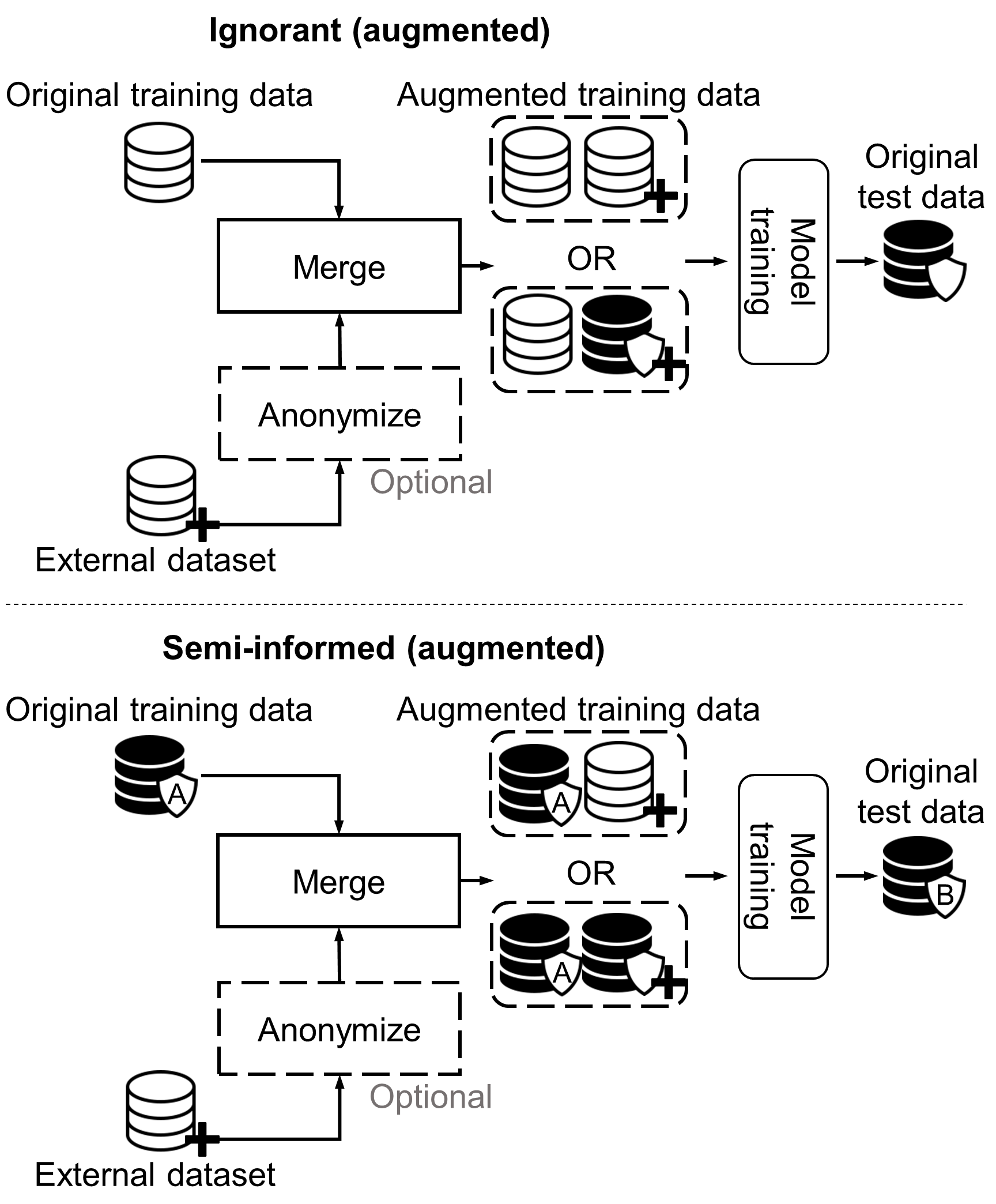}
  \caption{Data augmentation schemes in (top) scenario B (\textit{ignorant}) and (bottom) scenario C (\textit{semi-informed}).}
  \label{aug}
\end{figure}

Data augmentation has been widely used in speech applications based on deep neural networks to improve accuracy, especially under mismatched train-test data distributions. One of the approaches to increase model generalizability is to use external data augmentation, which refers to the case where data from external datasets curated for similar tasks are pooled with the in-domain data to increase training sample size \cite{deng2021improving, shahnawazuddin2020domain}. In our case, we aim to improve the generalizability of diagnostic models to samples anonymized by unknown algorithms. We propose to combine anonymized external data with the original data as an augmentation approach to mitigate the degradation caused by anonymization. We focus on two cases, namely augmenting the ignorant and semi-informed scenarios, in which we observed diagnostic models performed the worst. As shown in Figure~\ref{aug}, we experimented with four versions of the augmentation data, including the clean version (i.e., not anonymized), the McAdams-anonymized version, and the two GAN-anonymized versions. For simplicity, we sampled training and test data only from DiCOVA2 and used one of the other two datasets as external data. Since the performance of SVM and PCA-SVM were highly correlated, here we report only the improvement achieved with the openSMILE+SVM, MSR+SVM, and the LogMelSpec+BiLSTM diagnostic systems.

%

\subsection{Training and Inference Strategies}
\subsubsection{Training} For the systems that rely on hand-crafted features, training data normalization was achieved by removing the mean and scaling to unit variance. The fitted scaler was then applied to the validation and test data. Hyper-parameters were tuned on the held-out validation set. The optimal SVM regularization parameter was searched between $1e^{-5}$ and 1; the SVM kernel was set to linear; and the number of PCs was experimented from 100 to 300. To train the BiLSTM classifier, in turn, recordings were first zero-padded to 10-second length to ensure a fixed shape for logmelspec input; the spectrogram was then mean-variance normalized. Each mini-batch was composed of 64 samples with random shuffling, forced to contain both COVID-positive and COVID-negative samples. Unlike \cite{sharma2022second}, no oversampling of minority class or any other data augmentation techniques were used, as their effect on anonymization has yet to be quantified. The following hyper-parameters were used for training: Binary Cross Entropy (BCE) loss; Adam optimizer with an initial learning rate of $1e^{-4}$ and ${\textit{l}_{2}}$ regularization set to $1e^{-4}$. During the validation phase, an initial patience factor was set to 5 and reduced by 1 if the validation score did not increase. Training stopped whenever the patience factor reduced to 0, the number of training epochs was saved for the inference phase.
\subsubsection{Inference} For the first four systems, the pre-trained model with the highest validation score was then used for testing. As the BiLSTM classifier is more data-hungry, the optimal hyper-parameters found in the training phase were then used to train the classifier from scratch with the aggregated training and validation data. The number of training epochs maintained the same as that saved in the training phase.

\subsection{Evaluation Metrics}
Since all three datasets are imbalanced, the area under the receiver-operating-characteristic curve (AUC-ROC) was chosen as the primary metric to measure the diagnostics accuracy. We further calculated the 95\% confidence intervals (CIs) using 1000$\times$ bootstrap with replacement on the test set. According to \cite{platt2000bootstrap}, CIs can reflect the variability of diagnostics accuracy when the model is applied to a different population.

As mentioned previously, cosine similarity and misclassification rate were used to quantify the effectiveness of the three anonymization methods. The similarity scores are averaged across samples from all three datasets per method, where 0 represents no resemblance and 1 represents a perfect match between two tested speech conditions. While the Equal Error Rate (EER) is commonly used to evaluate anonymization efficacy, ground-truth speaker identifiers are required for each recording in order to verify if samples are from the same or different speakers. However, speaker identifiers were not available for the CSS and DiCOVA2 datasets. Instead, we rely on misclassification rate by employing a pre-trained speaker verification model from \cite{ravanelli2021speechbrain}. For each recording, the model outputs a binary decision (yes/no) if it believes a pair of anonymized and clean speech signals come from the same speaker. The misclassification rate is then calculated by dividing the number of misclassified pairs over the total number of pairs per method, which reflects the percentage of successfully anonymized recordings. For an ideal anonymization system, the misclassification rate is expected to be 100\%, i.e., the model should decide that all anonymized signals do not come from the same speaker as the clean signal counterpart. Lastly, computation time was recorded for each anonymization method, including the loading and exporting of the audio files. In the case of the two GAN methods, the loading time of the model itself was not taken into account in this computation.

\section{Experimental Results and Discussion}
\subsection{Task-1: Anonymization Results}
\begin{figure}
\centering
  \includegraphics[width=0.85\linewidth]{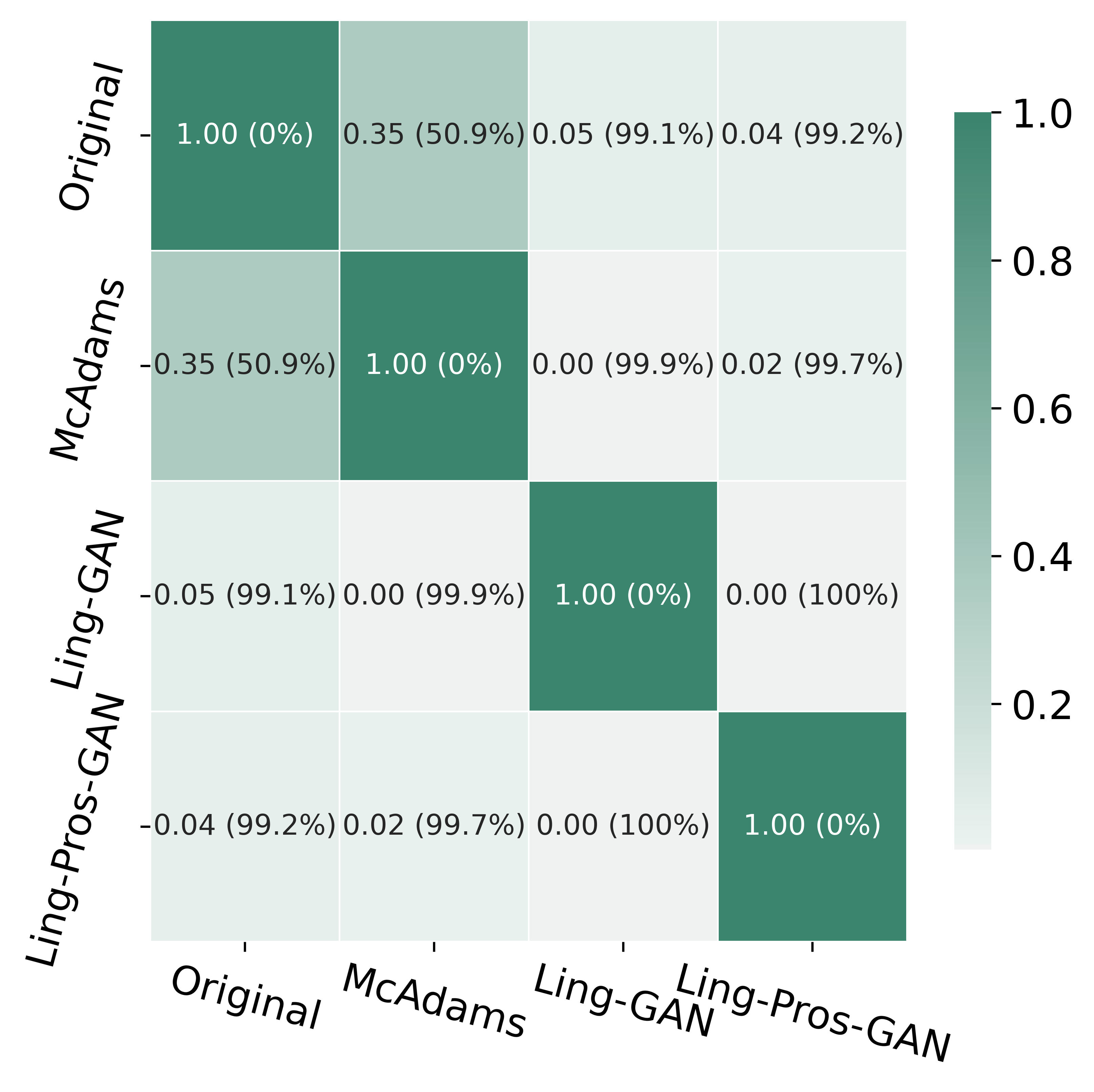}
  \caption{Cosine similarity between speech signals under different anonymization conditions averaged across three datasets. Values in the parentheses are the corresponding misclassification rates.}
  \label{heatmap}
\end{figure}

The average cosine similarity scores between the speaker embeddings of speech files anonymized by the different methods together with the misclassification rates are shown in Figure~\ref{heatmap}. As can be seen, near perfect anonymization performance was achieved with both GAN-based methods (misclassification rates), with almost no similarity with either the original speech or the speech anonymized by other methods. On the other hand, nearly half of the McAdams-anonymized samples can be successfully detected, suggesting some speaker-unique information still remained.


The computational complexity of the three anonymization methods is presented in Table \ref{tab:my_label} for all three datasets. While the GAN-based methods are shown to provide better anonymization effectiveness, it requires computational times approximately 10-20 times longer than using the McAdams coefficient method. The longest time was seen with Ling-Pros-GAN, since it requires extra time to extract prosody, which involves an online training loop, and to generate and find embeddings in real-time. As model loading time was not taken into account, the computation footprint of the GAN-based methods could be larger in real-world settings. Additionally, the GAN-based methods rely on several pre-trained neural networks with millions of parameters (e.g., 22.3 million for ECAPA-TDNN embedding extractor; 10 million for the generator), which could make it challenging to be deployed on mobile devices.

\begin{table}
\caption{Average computation time per speech file (second) with standard deviations using different anonymization methods for the three datasets.}
\centering
    \begin{tabular}{ccccc}
    \hline
    Method & CSS & DiC & Cam & Ave \\
    \hline
    McAdams Coef & 0.87\textpm0.10 & 1.15\textpm0.13 & 0.88\textpm0.92 & 0.97 \\
    Ling-GAN & 8.52\textpm2.93 & 10.22\textpm3.56 & 9.58\textpm2.70& 9.44 \\
    Ling-Pros-GAN & 26.49\textpm20.53 & 24.9\textpm11.61& 19.47\textpm11.61 & 23.62 \\
    \hline
    \end{tabular}
\label{tab:my_label}
\end{table}

\subsection{Task-2: Within-dataset Performance}
\begin{figure*}
  \centering
  \includegraphics[width=\linewidth]{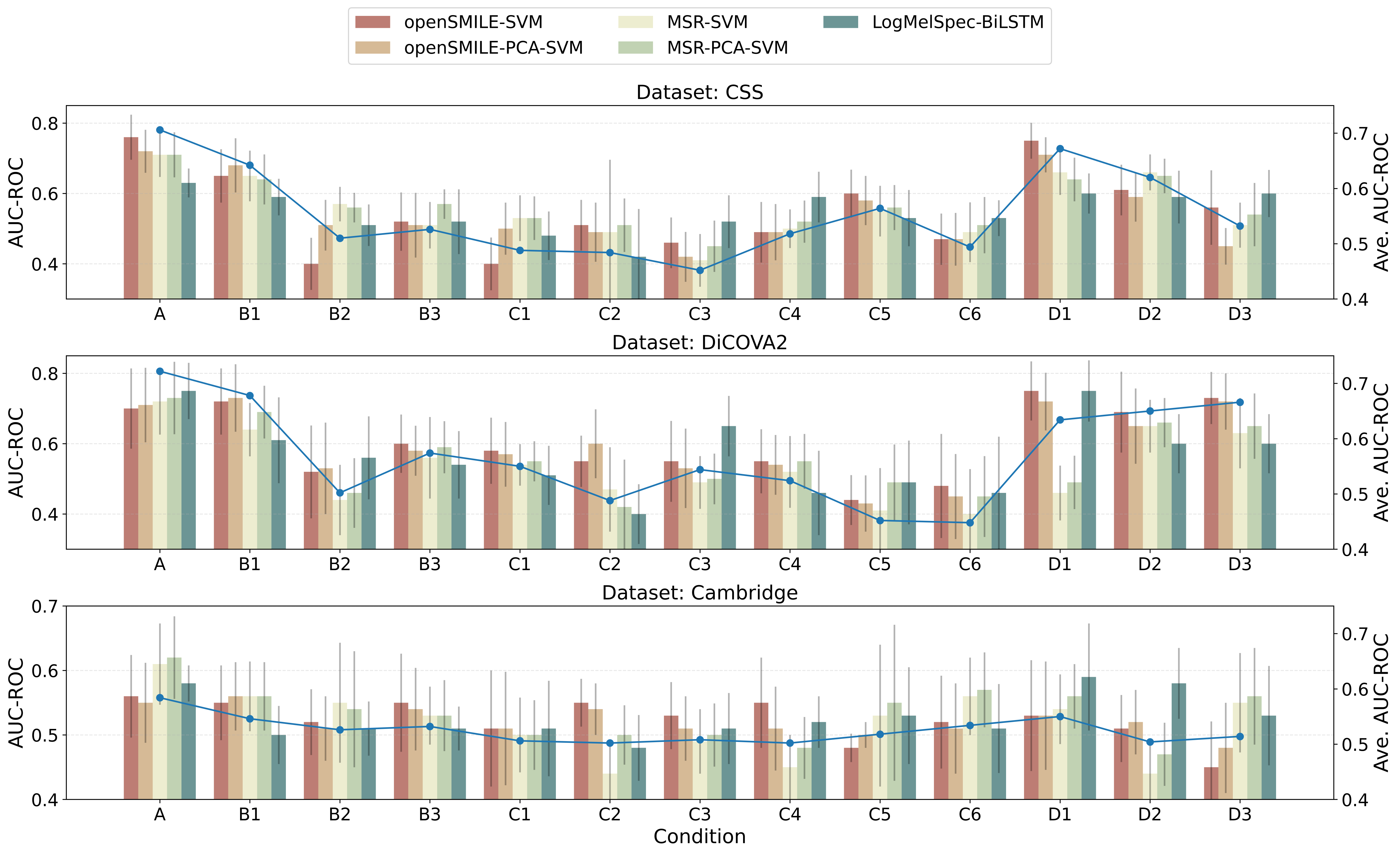}
  \caption{Within-dataset performance under different anonymization scenarios. Error bars represent the 95\% CIs. The line plot values correspond to the average AUC-ROC scores over the five diagnostic systems calculated per scenario.}
  \label{within}
\end{figure*}

The within-dataset performance of the five diagnostics systems under different anonymization scenarios is demonstrated in Figure~\ref{within}. As can be seen from the average AUC-ROC scores per scenario, the highest performance is achieved under scenario A, i.e., when anonymization is not performed. When the test data are anonymized using the McAdams coefficient (scenario B1), the average AUC-ROC score over all systems dropped by 8.9\% (CSS), 5.9\% (DiCOVA2), and 6.3\% (Cambridge) relative to scenario A. A substantial decrease was observed when using both the Ling-GAN and Ling-Pros-GAN anonymizers (scenario B2 and B3), where an average relative drop 22.5\% and 18.1\% was achieved respectively. Moreover, nearly all systems degraded to chance levels under scenario C where models were trained with data anonymized by one method and tested with data anonymized with another, suggesting that anonymization may drastically remove COVID-19 speech information. Diagnostic performance in the fully-informed scenarios is shown to be close to scenario A. Among the three anonymizers, McAdams anonymization leads to higher diagnostic performance on average in scenario D. Compared to the Ling-GAN, Ling-Pros-GAN shows higher performance on the English datasets (DiCOVA2 and Cambridge) and lower performance on the multilingual one (CSS).

Next, we evaluate the sensitivity of different diagnostics systems to anonymization and explore the relative drop in accuracy from scenario A to scenario B. Table~\ref{tab:ano} reports the average drops seen per dataset. As can be seen, the two GAN-based methods resulted in a substantially higher degradation relative to the McAdams coefficient method, with the Ling-GAN leading to the most severe decrease. This was expected and corroborates Task-1 results, where speaker embeddings of the GAN-anonymized speech showed practically no similarity to the original speech. Meanwhile, since Ling-Pros-GAN leaves the prosody intact and generates more COVID-like embeddings, it is likely to preserve more COVID-19 attributes than the Ling-GAN, thus rendering higher anonymized diagnostic performance. Previous studies have shown that speaker embeddings (e.g., x-vector) also contain other nonverbal information and can be used for speech para-linguistic tasks \cite{raj2019probing,van2021measuring}, such as speech emotion recognition \cite{pappagari2020x} and disease detection \cite{moro2020using, pappagari2020using}. While the GAN-based anonymizers substitute the original speaker embedding with a dissimilar speaker embedding, the obtained results suggest that health-related vocal characteristics are likely also discarded, thus resulting in significant drops in diagnostics accuracy.

\begin{table}[]
\caption{Drop in within-dataset AUC-ROC (\%) from scenario A to scenario B for different anonymization methods.}
\centering
    \begin{tabular}{>{\centering\arraybackslash}p{3cm}>{\centering\arraybackslash}p{0.7cm}>{\centering\arraybackslash}p{0.7cm}>{\centering\arraybackslash}p{0.7cm}>{\centering\arraybackslash}p{0.7cm}}
    \hline
    Anonymization method & CSS & DiC & Cam & Ave\\ \hline
    McAdams coefficient & 8.9 & 5.9 & 6.3 & 7.0 \\
    Ling-GAN & 27.3 & 30.5 & 9.8 & 22.5 \\
    Ling-Pros-GAN & 25.2 & 20.4 & 8.7 & 18.1 \\
    \hline
    \end{tabular}
\label{tab:ano}
\end{table}

Lastly, we use scenario A as the baseline and calculate the average drop in accuracy for scenario C, showing the impact that training models completely on anonymized data would have. For both openSMILE and MSR methods, we use the PCA-SVM pipeline to avoid the effects of difference in the number of features. The comparative results are reported in Table~\ref{tab:sensitivity}. As can be seen, all three diagnostic systems show degradaded performance, with the logmelspec+BiLSTM system shown to be on average more robust (21.6\%) to the semi-informed anonymization scenario. Notwithstanding, it should be highlighted that the logmelspec+BiLSTM system achieved the lowest AUC-ROC in scenario A. Interestingly, with the CSS dataset, the diagnostic system based on a BiLSTM and log-mel spectrogram input resulted in substantially lower degradation percentage compared to the two other systems based on traditional engineered features and classifiers. CSS is a multilingual dataset, thus hand-crafted features (e.g., syllabic rate, speech production features) used in these models may show more sensitivity to language.

\begin{table}[]
\caption{Drop in within-dataset AUC-ROC (\%) from scenario A to the average of all sub-conditions under scenario C for different diagnostics systems.}
\centering
    \begin{tabular}{>{\centering\arraybackslash}p{3cm}>{\centering\arraybackslash}p{0.7cm}>{\centering\arraybackslash}p{0.7cm}>{\centering\arraybackslash}p{0.7cm}>{\centering\arraybackslash}p{0.7cm}}
    \hline
    Diagnostics system & CSS & DiC & Cam & Ave \\ \hline
    openSMILE+PCA-SVM & 31.7 & 26.8 & 6.7 & 21.7\\
    MSR+PCA-SVM & 27.7 & 32.4 & 16.7 & 25.6\\
    LogMelSpec+BiLSTM & 18.8 & 34.0 & 12.1 & 21.6\\
    \hline
    \end{tabular}
\label{tab:sensitivity}
\end{table}

\subsection{Task-2: Cross-dataset Performance}
\begin{figure*}
\centering
  \includegraphics[width=\linewidth]{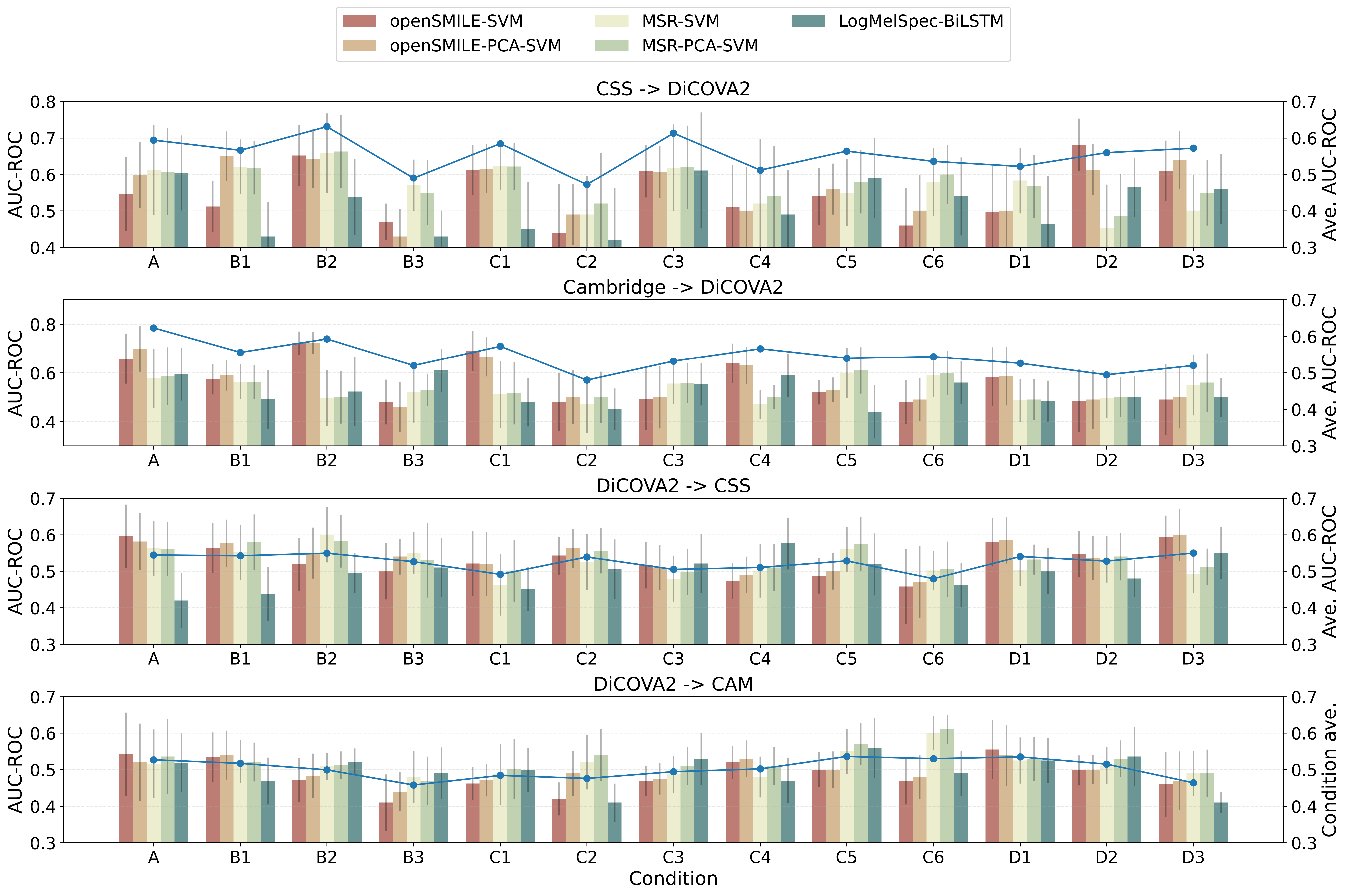}
  \caption{Cross-dataset performance under different anonymization scenarios. Error bars represent the 95\% CIs. The line plot values correspond to the average AUC-ROC scores over the five diagnostic systems calculated per scenario.}
  \label{cross}
\end{figure*}

\begin{figure}
    \centering
    \includegraphics[width=\linewidth]{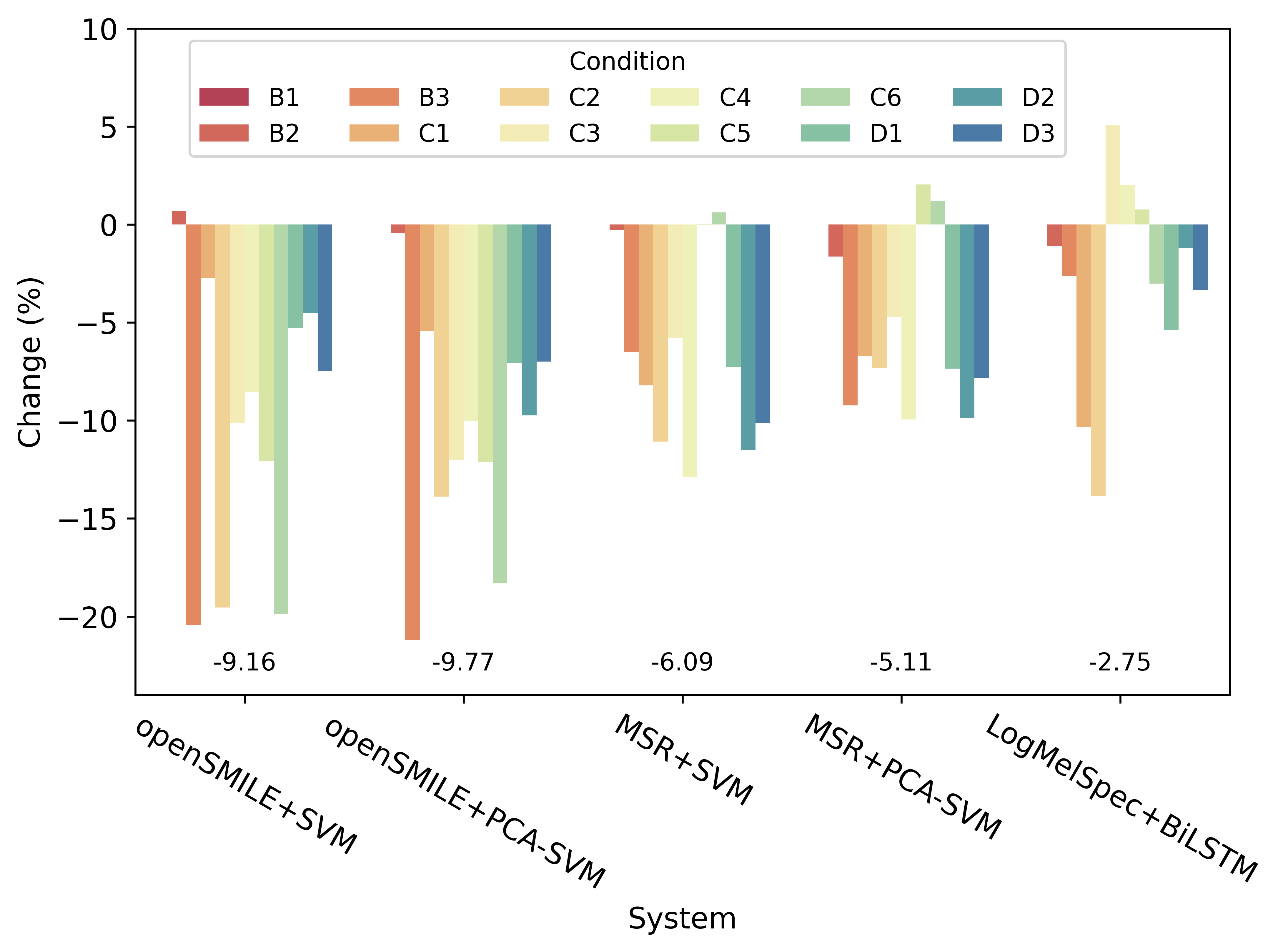}
    \caption{Relative changes in the AUC-ROC under different anonymization scenarios for all diagnostics systems in the cross-dataset experiment.}
    \label{fig:cross_change}
\end{figure}

Figure~\ref{cross} shows the cross-dataset performance under the thirteen different testing scenarios. In line with previous studies \cite{zhu2022generalizable, coppock2021end}, all five diagnostics systems demonstrated significantly lower performance relative to within-dataset results; the logmelspec+BiLSTM achieved the greatest drop in performance. Interestingly, in a few scenarios anonymization helped systems become more generalizable relative to the unprotected setting (e.g., scenarios B2 and C3 for the CSS-DiCOVA2 cross-database experiment). Figure~\ref{fig:cross_change} depicts the average change in accuracy relative to scenario A for all scenarios and diagnostic systems. While on average a 6.6\% drop in accuracy was seen across all five systems, an increase of 2\% and 5\% was achieved with MSR+SVM and logmelspec+BiLSTM systems for scenarios C4 and C2, respectively. It is important to note that both scenarios involved GAN-based anonymized test data, thus had typically the lower cross-dataset results to start off with.


%

\subsection{Explaining the degradation caused by different anonymizers}
While our study shows that typical anonymization systems lead to degraded diagnostic performance, it is unclear why different systems caused different levels of degradation and why some diagnostic models could still perform decently after anonymization. To answer these questions, we performed a comprehensive evaluation of the impact of different speech aspects on diagnostic performance, including the linguistic content, speaker representation, and prosody. Similar to the experimental setup of Task-1, we now compare the within-dataset performance obtained by three categories of speech features, namely (1) the phoneme-level features, including the number of mispronunciations (as opposed to the speech script), number of pauses, and number of phonemes uttered per second; (2) the speaker representation extracted by concatenating the pre-trained x-vector and ECAPA-TDNN embeddings~\cite{desplanques2020ecapa}; and (3) prosodic features, such as the low-level descriptors of the F0 contour. 

A Linear Discriminant Analysis (LDA) classifier is applied on top of each of the feature sets for classification. The results achieved by these features are reported in Table~\ref{tab:aspects}. Among the three feature sets, speaker embeddings appear to be the most crucial features for all datasets, corroborating with Task-1 results where the GANs suffered the most severe degradation, where the original speaker embeddings were entirely substituted. Such finding also suggests that speaker-unique attributes and health-related information are highly entangled in the speaker embeddings. Considering that existing anonymization systems rely heavily on these off-the-shelf speaker embeddings, it remains challenging to preserve the health information while altering only the speaker identifier.

While a group of studies reported prosody as a key biomarker to characterize speech disorders, such as dysarthria \cite{vyas2016automatic,kadi2013discriminative,ramos2020acoustic}, our results show that phoneme-level linguistic features outperform prosodic features for COVID-19 detection. Specifically, we found the number of pauses and number of mispronunciations to be the most important phoneme-level features, with COVID-positive samples demonstrating more mispronunciations and fewer pauses. While the correlation between phoneme-level features and COVID-19 status has not been systematically studied, similar features have been examined for other diseases affecting speech production. For example, \cite{darling2020impact} shows that individuals with Parkinson's disease produced fewer pauses at syntactic boundaries; the statistics of pauses have been shown crucial for diagnosing neuromuscular disorders, such as dysarthria~\cite{noffs2018speech}. Since GAN-based systems left linguistic content intact during anonymization, these findings help explain why the diagnostic models could perform above chance-level even when only the phoneme sequences were preserved during anonymization.

\begin{table}
\caption{Diagnostic performance achieved by different categories of speech features}
    \centering
    \begin{tabular}{ccccc}
    \toprule
    \multirow{2}{*}{Feature} & \multicolumn{3}{c}{AUC-ROC} \\
    \cmidrule{2-4}
    & CSS & DiCOVA2 & Cambridge\\
    \midrule
    Linguistic & .561 & .632 & .555\\
    Speaker & \bf{.739} & \bf{.697} & \bf{.571}\\
    Prosodic & .541 & .564 & .520 \\
    \bottomrule
    \end{tabular}
\label{tab:aspects}
\end{table}

\subsection{Visualizing speech processed by different anonymizers}
\begin{figure}
\centering
\includegraphics[width=\linewidth]{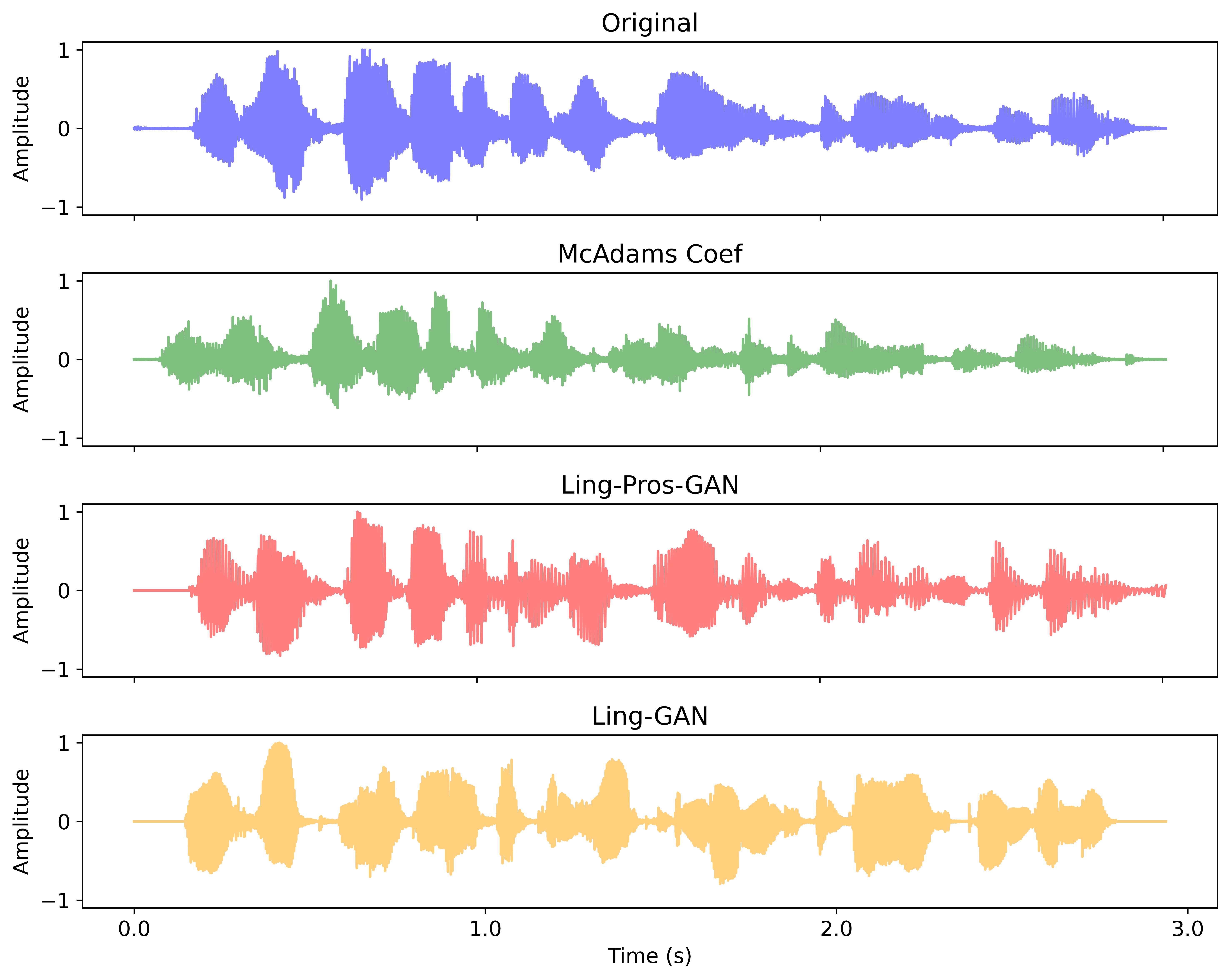}
\caption{A comparison of the waveforms processed by the three anonymizers and the original speech.}
\label{fig:signal}
\end{figure}

To better understand the impact of different anonymization methods on speech characteristics, we first visualize the waveform of the speech processed by the three anonymizers (see Fig.~\ref{fig:signal}) for a direct comparison. As can be seen, those processed by the McAdams anonymizer and Ling-Pros-GAN share higher similarities in the waveform envelope shape with the original signal compared to the one generated by Ling-GAN. The difference seen in the plot is in line with the architecture design of different anonymizers. Among the three, Ling-GAN loses prosody and most of the speaker attributes, hence is expected to cause the highest amount of changes in the anonymized speech. The Ling-Pros-GAN and McAdams anonymizer, in turn, leave the speech rhythm untouched (i.e., duration and energy of phonemes), hence leading to higher resemblance in the waveform envelope.

\begin{figure}
\centering
\subfloat[]{\includegraphics[width=\linewidth]{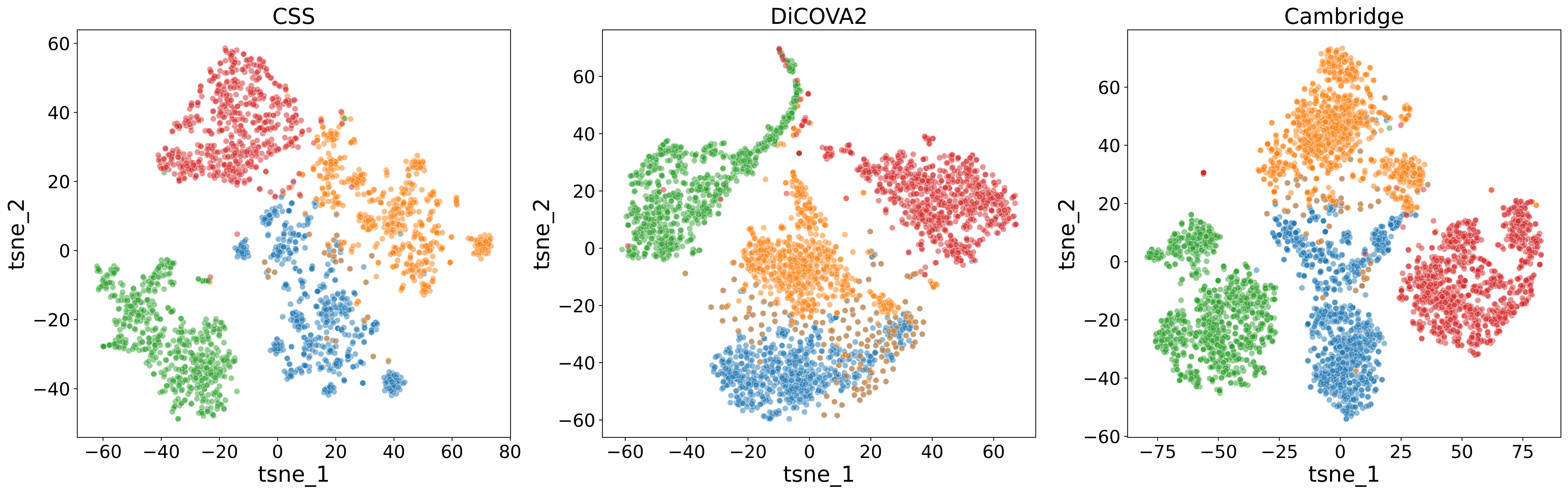}%
\label{kde_opensmile}}

\subfloat[]{\includegraphics[width=\linewidth]{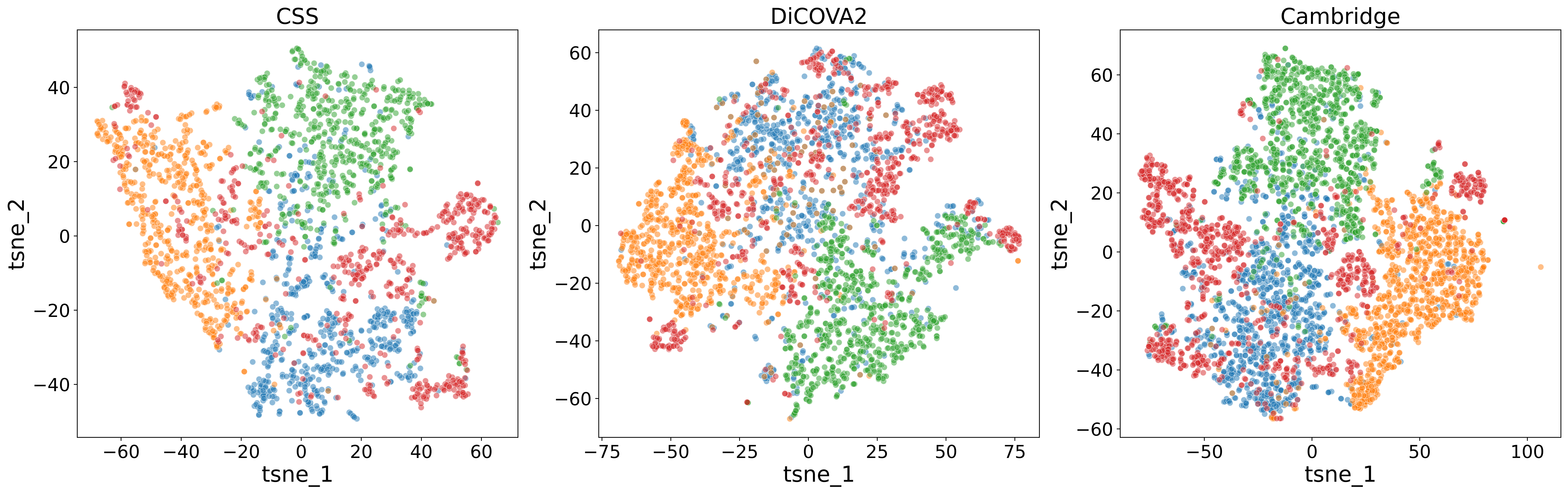}%
\label{kde_msr}}

\subfloat[]{\includegraphics[width=\linewidth]{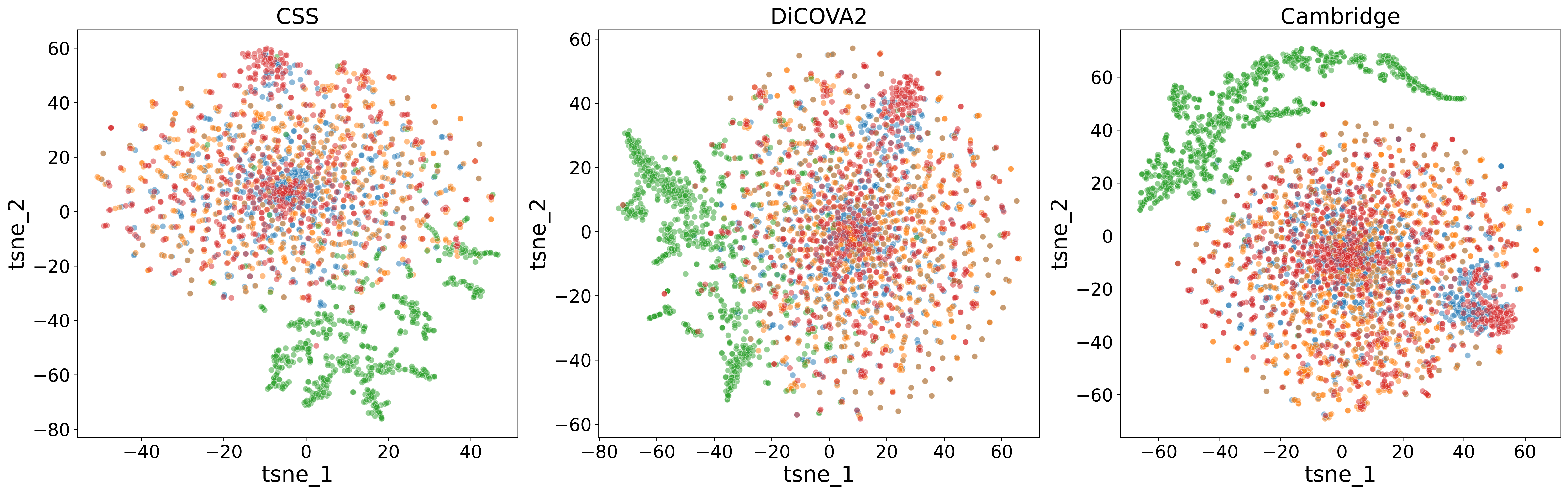}%
\label{kde_logmelspec}}

\caption{t-SNE clusters of anonymized speech features for different feature sets, namely: (a) openSMILE, (b) MSR, and (c) logmelspec. Blue dots corresponds to original speech; orange to McAdams coefficient anonymized speech; red to Ling-Pros-GAN anonymized speech; and green to Ling-GAN anonymized speech.}
\label{kde}
\end{figure}

Next, t-SNE plots are used to visualize the distribution of the speech features in two dimensions. Figure~\ref{kde} shows the clusters of speech anonymized with different methods (computed from the training and validation data) and for the three features modalities explored herein: openSMILE (subplots a), MSR (b), and logmelspec (c). As can be seen, for all three feature sets, the distribution of clean speech (blue) is closer to that of the McAdams anonymized speech (orange) and Ling-Pros-GAN anonymized speech (red), while the Ling-GAN anonymized speech (green) shows the least similarity with the other two, corroborating findings from Tasks 1 and 2. 

Moreover, it can be seen from Figure~\ref{kde_opensmile} and Figure~\ref{kde_msr} that the clusters computed from openSMILE and MSR features show little overlap, while clusters of the logmelspec features show great overlap (Figure~\ref{kde_logmelspec}). Together with Task-2 results, this shift in the feature space is likely the main cause of the higher decrease observed in the openSMILE and MSR systems under different anonymization settings. Meanwhile, since all anonymization methods keep the speech content intact and change only the nonverbal attributes, a greater shift in feature space may indicate a stronger correlation with the para-linguistic aspect and less with the linguistic aspect. This echoes with previous studies which showed that openSMILE and MSR features are preferred over logmelspec features in characterizing emotional and unnatural speech \cite{falk2012characterization, eyben2015real}.

\subsection{Task-3: Improving Diagnostics Performance with Data Augmentation}
Lastly, we investigate the impact of using anonymized external data for data augmentation and see its impact on the performance achieved with scenarios B and C. With scenario C, we chose sub-condition C1 and C3. To quantify the relative improvement, we used the within-dataset performance achieved in scenario A as the baseline and calculated the amount of performance increase seen (in percentage). The relative changes observed with the three diagnostics systems are reported in Table~\ref{tab:task3}. Here, we explore augmentation with two different datasets and with four different methods: original, McAdams, Ling-GAN, and Ling-Pros-GAN. As can be seen, when test data are anonymized using the McAdams coefficient (B1), the highest improvement is generally achieved when the diagnostic system is augmented with the original data. In turn, when the test data are anonymized using the GAN-based method (B2), augmenting the set with GAN-anonymized data from another dataset leads to a higher increase. Similar results are shown in scenarios C1 and C2, where clean and GAN-anonymized augment data result in more significant improvements. While not the top-performer, the McAdams method is shown to be a reliable augmentation strategy, especially for the openSMILE features. Overall, these findings suggest that anonymization has the potential to be used as a data augmentation approach to improve COVID-19 diagnostics accuracy when tested on anonymized data. 
\begin{table}[]
\caption{Change of AUC-ROC scores achieved in scenarios B and C after data augmentation (given in \%). Bold values indicate the highest improvement with each diagnostics system under a given scenario.}
    \begin{tabularx}{\linewidth}{c@{\hspace{1.3\tabcolsep}}c@{\hspace{1.5\tabcolsep}}c@{\hspace{1.3\tabcolsep}}ccc}
    \toprule
    \multirow{2}{*}{Scen.} & \multirow{2}{*}{Aug. data} & \multirow{2}{*}{Augment} & \multicolumn{3}{c}{Diagnostics System} \\
    \cmidrule(lr){4-6}
    & & & openSMILE & MSR& logmelspec \\
    \midrule
    \multirow{8}{*}{B1} & \multirow{4}{*}{CSS} & Clean & \bf{0.6} & \bf{9.9} & -30.6 \\
    & & McAdams & -5.0 & 2.7 & -23.7 \\
    & & Ling-GAN & -2.4 & 5.2 & -7.9 \\
    & & Ling-Pros-GAN & -5.0 & -9.5  & -8.9 \\
    \cmidrule(lr){3-6}
    & \multirow{4}{*}{Cam} & Clean & -1.5 & -6.4 & -9.7 \\
    & & McAdams & -12.1 & -23.7 & 0.1 \\
    & & Ling-GAN & -14.1 & -2.2 & \bf{5.7} \\
    & & Ling-Pros-GAN & -1.2 & -10.7 & -1.1 \\
    \midrule
    \multirow{8}{*}{B2} & \multirow{4}{*}{CSS} & Clean & 21.3 & 3.9 & -7.9 \\
    & & McAdams & 3.8 & 15.0 & -15.7 \\
    & & Ling-GAN & \bf{24.5} & 25.5 & -1.6 \\
    & & Ling-Pros-GAN & 4.1 & 4.9 & -5.3\\
    \cmidrule(lr){3-6}
    & \multirow{4}{*}{Cam} & Clean & 21.5 & \bf{31.0} & -7.9 \\
    & & McAdams & 4.8 & 16.6 & 1.8 \\
    & & Ling-GAN & 16.1 & 23.2 & \bf{18.6} \\
    & & Ling-Pros-GAN & -5.4 & 4.6 & 3.9\\
    \midrule
    \multirow{8}{*}{C1} & \multirow{4}{*}{CSS} & Clean & 14.4 & -8.8 & \bf{11.7}\\
    & & McAdams & -2.2 & -6.7 & -6.6\\
    & & GAN & 16.6 & -0.4 & 8.2\\
    & & Ling-Pros-GAN & 0.0 & \bf{9.5} & 3.8 \\
    \cmidrule(lr){3-6}
    & \multirow{4}{*}{Cam} & Clean & \bf{18.7} & 5.0 & -0.8\\
    & & McAdams & 1.9 & -30.5 & -9.6\\
    & & Ling-GAN & 7.4 & -10.1 & 2.9\\
    & & Ling-Pros-GAN & -21.0 & -14.0 & 1.3\\
    \midrule
    \multirow{8}{*}{C3} & \multirow{4}{*}{CSS} & Clean & 5.9 & 11.1 & -28.3\\
    & & McAdams & 5.3 & -13.9 & -10.8\\
    & & Ling-GAN & 2.0 & \bf{26.2} & -5.4 \\
    & & Ling-Pros-GAN & 14.5 & -5.4 & -3.7 \\
    \cmidrule(lr){3-6}
    & \multirow{4}{*}{Cam} & Clean & 15.7& -8.7 & -8.5\\
    & & McAdams & 15.5 & -0.4 & -16.6\\
    & & Ling-GAN & 1.5 & -20.7 & \bf{5.6}\\
    & & Ling-Pros-GAN & \bf{22.0} & -4.7 & 2.1\\

    \bottomrule
    \end{tabularx}
\label{tab:task3}
\end{table}

\subsection{Limitations, Biases, and Future Work}
The study’s principal aim was to validate the effectiveness of anonymization methods within and across datasets in the context of assessing voice-based COVID-19 diagnostic accuracy. While we investigated three anonymization methods, other methods are emerging continuously (e.g., \cite{deng2022v, miao2023speaker}); thus, the findings reported herein should be validated with more recent methods. In the present study, the ASR anonymization method developed on English speech was applied to the multilingual CSS dataset. The finding that GAN-based anonymization had the lowest cross-dataset performance results may suggest challenges in applying this method in multilingual datasets and non-English speaking populations. In the future, multilingual GANs should be explored to avoid unfair outcomes \cite{morley2020ethics} due to certain languages or cultural settings being excluded from the training and testing datasets. Moreover, while the injection of anonymized external data showed to be a useful data augmentation strategy, the final results were still at times lower than those achieved in the classical ``unprotected'' setting. This suggests that health-related information is being discarded during the anonymization process, thus future work could explore the development of diagnostic-aware anonymization methods that keep such discriminatory information intact.

Beyond tackling these limitations mentioned above, future work into voice-based diagnostics should be mindful of potential biases during data collection that could lead to confounds for both the anonymization and diagnostic steps. These confounders, if not properly dealt with, can reinforce the systemic nature of biases, for instance in relation to gender and racioethnic groups, that already exist within the healthcare system, thus transferring them to automated diagnostic systems. While \cite{han2022sounds} already showed some impact of sampling rate on diagnostic accuracy, several other potential biases may exist at the methodological level. For example, sociodemographic biases may emerge if age is not taken into consideration, as cognitive limitations (e.g., difficulty in speech planning or lexical access) associated with aging could alter speech patterns that could affect overall diagnostic accuracy. Recent work has shown that socioeconomic status could serve as a bias in COVID-19 detection \cite{zhu2023bias}. For example, as data was collected from participants from home, those living in crowded conditions could have resulted in increased background noise levels that negatively affected anonymization and diagnostic efficacy. Moreover, disadvantaged populations have been shown to have more chronic respiratory diseases \cite{pleasants2016defining} and higher levels of mood disorders and psychological distress \cite{drapeau2012epidemiology}. As such, anonymization processes affecting para-linguistic features associated with depressed mood may disproportionately affect those with a low socioeconomic status. Addressing biases in automated voice anonymization and diagnosis systems is beyond the scope of this paper and is left for future studies.

\section{Conclusion}
In this study, we comprehensively evaluated the impact of three voice anonymization methods on the accuracy of five leading COVID-19 detection systems as well as the anonymization efficacy. All anonymization methods showed to degrade diagnostics accuracy, where the most severe degradation was seen with the systems that directly altered speaker embeddings. Our findings suggest that existing methods lack the capability of effectively preserving diagnostic information while obfuscating speaker identifiers. Lastly, we explored the use of anonymized external data as a data augmentation tool and promising results were obtained.

\section*{Acknowledgments}
The authors would like to thank the developers of the CSS, DiCOVA2, and Cambridge datasets for making them available to the community for research purposes. The developers of the datasets do not bear any responsibility for the analysis and results presented in this paper. All results and interpretations only represent the view of the authors. The authors acknowledge funding from INRS and NSERC.

\bibliographystyle{IEEEtran}
\bibliography{ref}

\vfill

\end{document}